\pdfoutput=1

\documentclass[11pt]{article}

\usepackage[final]{acl}

\usepackage{times}
\usepackage{latexsym}
\usepackage{amsmath,bm}
\usepackage{amssymb}
\usepackage{bbm}
\usepackage{multirow}
\usepackage{tikz}
\usepackage{pgfplots}
\usetikzlibrary{calc,patterns,angles,quotes}
\usepackage{subcaption}
\usepackage{placeins}
\usepackage{xcolor}
\usepackage{ifthen}
\usepackage{relsize}
\usepackage{subcaption}

\usepackage[T1]{fontenc}

\usepackage[utf8]{inputenc}

\usepackage{microtype}

\usepackage{inconsolata}

\usepackage{graphicx}

\title{Revisiting Intermediate-Layer Matching in Knowledge Distillation:\\
Layer-Selection Strategy Doesn't Matter (Much)}

\author{
 \textbf{Zony Yu},\quad
 \textbf{Yuqiao Wen,\quad
 \textbf{Lili Mou}$^{*}$
 }\\
 \fontsize{10pt}{12pt}\selectfont
 Dept. Computing Science \& Alberta Machine Intelligence Institute (Amii), University of Alberta \\
 \fontsize{10pt}{12pt}\selectfont
 $^*$Canada CIFAR AI Chair, Amii \\
 \href{mailto:zony249@gmail.com}{\fontsize{10pt}{12pt}\selectfont \texttt{zony249@gmail.com}}, \ \ 
 \href{mailto:yq.when@gmail.com}{\fontsize{10pt}{12pt}\selectfont \texttt{yq.when@gmail.com}},
 \ \ \href{mailto:doublepower.mou@gmail.com}{\fontsize{10pt}{12pt}\selectfont \texttt{doublepower.mou@gmail.com}}
}

\newboolean{usecolor}
\setboolean{usecolor}{false}

\definecolor{bluish}{RGB}{76, 93, 120}
\DeclareRobustCommand{\mycolor}[2]{%
    \ifthenelse{\boolean{usecolor}}{\textcolor{#1}{#2}}{#2}%
}

\begin{document}
\maketitle
\begin{abstract}
Knowledge distillation (KD) is a popular method of transferring knowledge from a large ``teacher'' model to a small ``student'' model.  Previous work has explored various layer-selection strategies (e.g., forward matching and in-order random matching) for intermediate-layer matching in KD, where a student layer is forced to resemble a certain teacher layer. In this work, we revisit such layer-selection strategies and observe an intriguing phenomenon that layer-selection strategy does not matter (much) in intermediate-layer matching---even seemingly nonsensical matching strategies such as {\textit{reverse matching}} still result in surprisingly good student performance. We provide an interpretation for this phenomenon by examining the angles between teacher layers viewed from the student's perspective. Our work sheds light on KD practice, as layer-selection strategies may not be the main focus of KD system design, and vanilla forward matching works well in most setups.\footnote{The code is released at the following repo: \url{https://github.com/MANGA-UOFA/Layer-Selection}}
\end{abstract}

\section{Introduction}
\label{sec:intro}

Large language models have achieved impressive performance in various NLP tasks \citep{gpt4o, deepseek}. However, they need a large number of parameters, making the models cumbersome and difficult to run in resource-restricted scenarios. Knowledge distillation (KD; \citeauthor{hintonkd}, \citeyear{hintonkd}) is a widely adopted method to reduce model parameters by training a small ``student'' model from a large ``teacher.'' With KD, the student is often able to retain most of the teacher's performance while using a fraction of its parameters \citep{minillm,deepseek,wen-etal-2025-knowledge,qwen3}.


Common KD approaches can be generally divided into two categories: prediction matching and intermediate-layer matching. Matching the prediction is usually mandatory, as it informs the student of the task to solve. This can be achieved by minimizing the divergence of predicted distributions~\cite{hintonkd,fdistill,sinkhorn} or using reinforcement learning~\cite{hao2022teacher,llmr}.

Intermediate-layer matching distills the teacher's hidden states (i.e., intermediate layers) to the student~\cite{patientkd,tinybert,ineffectiveness}. This approach often involves minimizing the distance between the student's and teacher's hidden states (usually with a linear mapping if the dimensions do not match). Also, a layer-selection strategy is required to specify which teacher layer is matched to which student layer.

Traditionally, researchers have explored various layer-selection strategies.~\citet{patientkd} match the student's layers to evenly spaced teacher layers;  \citet{alpkd} and~\citet{universal} learn an attention mechanism over the teacher's layers; \citet{railkd} match the student's layers to randomly selected layers from the teacher, albeit in sorted order; and \citet{minilmv2} match the last student layer to a teacher layer close to the end. 


Despite numerous previous efforts, we observe that the effectiveness of existing layer-selection strategies is not convincing, as previous studies often lack proper controlled comparisons. For example,~\citet{alpkd} only compare their work with the vanilla layer-matching baseline, and~\citet{minilmv2} compare layer-selection strategies without controlling weight initializations. In addition, existing work tends to omit important distillation settings such as decoder-only model architectures.
As a result, there is a lack of understanding about how different strategies compare against each other.

In this work, we aim to better understand how
layer-selection strategies affect student performance in KD. We conduct systematic controlled
experiments across four models (BERT, BART, T5,
and Qwen3) on eight tasks (including both classification and generation). We also consider two
weight initialization strategies: random initialization
and weight copying from the teacher. Our experiments reveal intriguing findings that
\begin{enumerate}
\item The layer-selection strategy doesn't matter (much), as different strategies perform similarly to each other, and 
\item Regardless of layer-selection strategies, intermediate-layer matching itself is a highly effective KD method, outperforming the no-matching baseline.
\end{enumerate}
We provide an interpretation for this phenomenon based on geometric analysis of the hidden states: from the student's point of view, the angles between two teacher layers are often acute; thus, matching any teacher layer pulls the student layer in a similar direction. As a result, intermediate-layer matching indeed benefits KD, although the matching strategy does not matter (much). 

Our study offers a practical suggestion for KD: we recommend KD practitioners to focus on other aspects of KD systems (e.g., distillation temperature, choice of $f$-divergence loss functions), while simply using forward matching for intermediate-layer matching as a default strategy, if computing resources are limited.

\section{Background and Related Work} 





 


Knowledge Distillation (KD) is a method of transferring rich knowledge contained in a teacher model to a student model. To inform the student of the task, it is essential to match the student's and teacher's predictions~\citep{bucila}. \citet{hintonkd} suggest minimizing the Kullback--Leibler (KL) divergence between the teacher and student distributions, denoted by $p$ and $q_{\bm\theta_s}$, respectively. The KL distillation loss is given by
\begin{equation}
    \mathcal{L}_{\text{KL}}(\bm\theta_s) =  
    \mathbb{E}_{\mathrm y\sim p(\mathrm y| \mathrm x)}\Big[
        \log{
            \dfrac{p(\mathrm y | \mathrm x)}
            {q_{\bm\theta_s}(\mathrm y | \mathrm x)}
        }
        \Big]
\end{equation}
where $\mathrm x$ represents the input, and the output $\mathrm y$ (conditioned on $\mathrm x$) is sampled from $p$. The student's parameters $\bm\theta_s$ are optimized, whereas the teacher's parameters are frozen. 

Other than minimizing KL, different prediction matching approaches have been proposed.  When the teacher distribution is diverse, for example, the reverse KL divergence~\cite{engine,minillm} is used due to its mode-seeking behavior, i.e., the student only focuses on one of the high-probability regions in the teacher distribution~\cite{mode-seeking}. \citet{fdistill} propose an $f$-divergence KD framework, where symmetric divergences (such as Jensen--Shannon and total variation distance) provide a balance between mode averaging and mode seeking. Reinforcement learning can also be applied to KD~\cite{hao2022teacher,llmr}, which makes the student aware of its prefix and addresses the exposure bias problem~\cite{exp-bias}.

Regarding intermediate-layer matching, it distills the teacher's hidden states, thus providing additional supervisory signals to the student~\cite{patientkd, wen-etal-2025-knowledge}. Let $\mathcal M=\{(\varsigma_i, \tau_i)\}_{i}$ be the mapping between certain student and teacher layers, i.e., the $\varsigma_i$th layer of the student is mapped to the $\tau_i$th layer of the teacher. Intermediate-layer matching typically penalizes the distance between the matched layers, given by
\begin{equation}
    \mathcal L_{\text{hid}}( \bm\theta_s, \{\boldsymbol A_i\}_i) = \sum_{i} \operatorname{dist}(\boldsymbol A_i \bm h_{\varsigma_i}^{(s)}, \boldsymbol h_{\tau_i}^{(t)}) 
\end{equation}
where $\operatorname{dist}$ is a distance metric (such as mean squared error). The trainable linear operator $\bm A_i$ transforms the student's hidden state $\bm h_{\varsigma_i}^{(s)}$ to the space of the teacher's hidden state $\boldsymbol h_{\tau_i}^{(t)}$, if their dimensions do not match. Otherwise, $\bm A_i$ may be an identity matrix.

Intermediate-layer matching can be applied to different types of representations. Traditionally, this is achieved by matching the student's and teacher's activations~\cite{patientkd, distilBERT}. Other studies match attention logits~\cite{tinybert}, query--key--value relations~\cite{minilmv2}, and cross-sample relations~\cite{rkd, fcd}. In our work, we focus on matching activations because it is the most fundamental approach in intermediate-layer matching. 

Various layer-selection strategies have been proposed for matching a shallow student to a deep teacher. \citet{patientkd} and \citet{tinybert} suggest mapping evenly spaced teacher layers to the student. \citet{alpkd} and~\citet{universal} match each student layer to a weighted combination of all teacher layers to retain more knowledge. \citet{railkd} randomly reselect a sequence of teacher layers to match with the student (in order) after each epoch, so that the student is exposed to different teacher layers.~\citet{minilmv2} and~\citet{ilm-overfitting} suggest mapping the last non-prediction student layer to a teacher layer close to the end.


\begin{table*}[t!]
\centering

\resizebox{\textwidth}{!}{%
\begin{tabular}{|ll|l|r|cccc|cc|}
\hline
\multicolumn{2}{|l|}{} &  & \multicolumn{1}{c|}{} & \multicolumn{4}{c|}{Classification Tasks} & \multicolumn{2}{c|}{Generation Tasks} \\ \cline{5-10} 
\multicolumn{2}{|l|}{\multirow{-2}{*}{Model}} & \multirow{-2}{*}{\begin{tabular}[c]{@{}l@{}}Layer \\ Matching\end{tabular}} & \multicolumn{1}{c|}{\multirow{-2}{*}{\#}} & \begin{tabular}[c]{@{}c@{}}MNLI-m/mm\\ Acc\end{tabular} & \begin{tabular}[c]{@{}c@{}}QQP\\ Acc \ / \ F1\end{tabular} & \begin{tabular}[c]{@{}c@{}}QNLI\\ Acc\end{tabular} & \begin{tabular}[c]{@{}c@{}}SST-2\\ Acc\end{tabular} & \begin{tabular}[c]{@{}c@{}}DART\\ BLEU\end{tabular} & \begin{tabular}[c]{@{}c@{}}WMT16 \\ BLEU\end{tabular} \\ \hline
\multicolumn{1}{|l|}{} & Previous work & -- & 1 & 84.6 \ / \ 83.4 & \ \ \ --  \ \  \    / \ 71.2 & 90.5 & 93.5 & 48.56 & 25.82 \\
\multicolumn{1}{|l|}{\multirow{-2}{*}{Teacher}} & Our replication & -- & 2 & 84.5 \ / \ 84.1 & 89.0 \ / \ 71.4 & 90.8 & 93.1 & 48.80 & 25.90 \\ \hline
\multicolumn{1}{|l|}{} &  & None & 3 & 63.2 \ / \ 63.6 & 81.5 \ / \ 56.4 & 61.2 & 81.1 & {\color[HTML]{000000} \textbf{38.76}} & 8.02 \\
\multicolumn{1}{|l|}{} &  & Forward & 4 & 72.5 \ / \ 72.0 & 83.9 \ / \ 61.3 & 64.7 & 85.1 & {\color[HTML]{000000} 32.64} & \textbf{18.13} \\
\multicolumn{1}{|l|}{} &  & Reverse & 5 & 69.3 \ / \ 68.9 & \textbf{84.3} \ / \ \textbf{61.8} & \textbf{65.2} & 83.3 & {\color[HTML]{000000} 33.12} & 17.15 \\
\multicolumn{1}{|l|}{} &  & All-to-one & 6 & \textbf{74.0} \ / \ \textbf{73.8} & 83.4 \ / \ 60.2 & 65.0 & \textbf{85.4} & {\color[HTML]{000000} 33.86} & 17.16 \\
\multicolumn{1}{|l|}{} & \multirow{-5}{*}{\begin{tabular}[c]{@{}l@{}}Randomly \\ initialized\end{tabular}} & Out-of-order rand & 7 & 70.5 \ / \ 70.5 & 82.4 \ / \ 58.8 & 64.4 & 82.9 & {\color[HTML]{000000} 32.73} & 16.71 \\ \cline{2-10} 
\multicolumn{1}{|l|}{} &  & None & 8 & 77.4 \ / \ 76.5 & 87.6 \ / \ 67.1 & 81.2 & 88.7 & 46.32 & 22.36 \\
\multicolumn{1}{|l|}{} &  & Forward & 9 & \textbf{79.7} \ / \ \textbf{78.8} & \textbf{88.2} \ / \ \textbf{69.1} & \textbf{83.8} & \textbf{92.3} & 47.94 & \textbf{22.65} \\
\multicolumn{1}{|l|}{} &  & Reverse & 10 & 79.2 \ / \ 78.2 & 88.1 \ / \ 68.3 & 83.2 & 89.6 & \textbf{48.45} & 21.57 \\
\multicolumn{1}{|l|}{} &  & All-to-one & 11 & 79.4 \ / \ 78.7 & 87.6 \ / \ 68.6 & 82.8 & 91.4 & 47.10 & 21.89 \\
\multicolumn{1}{|l|}{\multirow{-10}{*}{Student}} & \multirow{-5}{*}{Weights copied} & Out-of-order rand & 12 & 79.0 \ / \ 78.0 & 87.5 \ / \ 67.2 & 82.6 & 90.7 & 47.99 & 22.01 \\ \hline
\end{tabular}%
}

\vspace{-.2cm}

\caption{Main results. We use BERT on classification tasks, BART on DART, and T5 on WMT16.}\vspace{-.2cm}
\label{tab:tab1}
\end{table*}

\begin{table}[t!]
\resizebox{1.03\columnwidth}{!}{
\hspace{-8px}
\mycolor{red}{
\begin{tabular}{|l|l|r|cc|}
\hline
Model & Layer Matching & \multicolumn{1}{c|}{\#} & \begin{tabular}[c]{@{}c@{}}HellaSwag\\ Acc\end{tabular} & \begin{tabular}[c]{@{}c@{}}CoQA\\ Exact Match\end{tabular} \\ \hline
Teacher & -- & 1 & 63.11 & 75.02 \\ \hline
\multirow{5}{*}{Student} & None & 2 & 35.16 & 33.73 \\
 & Forward & 3 & \textbf{37.85} & 35.40 \\
 & Reverse & 4 & 35.01 & \textbf{37.67} \\
 & All-to-one & 5 & 34.99 & 37.35 \\
 & Out-of-order rand & 6 & 35.47 & 34.80 \\ \hline
\end{tabular}
}
}
\caption{KD results on more challenging tasks using the recent Qwen3 model.}\vspace{-.2cm}
\label{tab:llm}
\end{table}



Overall, although previously proposed methods work well within their environments, it remains unclear how various layer-selection strategies compare under a controlled setup, or the extent they contribute to KD. Thus, we address this by conducting systematic investigations across numerous tasks, models, and initialization strategies, and provide an interpretation for our observations.

\section{Approaches}
\label{sec:setup}

Intermediate-layer matching requires a strategy to select which teacher layers are matched with which student layers. 
In this study, we explore both standard and seemingly nonsensical layer-selection strategies to uncover its effect on intermediate-layer matching KD.

\textbf{Forward Matching.} In this variant, lower student layers are matched to lower teacher layers. In particular, we follow \citet{patientkd} and select evenly spaced teacher layers for matching.

\textbf{All-to-One Matching.} In this variant, all student layers are matched to the middle teacher layer. While the idea of matching to one layer is proposed by previous studies~\cite{minilm,minilmv2}, we slightly modify their approaches (i.e., matching all student layers instead of one), for fair comparison with the rest of our settings. 

\textbf{Reverse Matching.} We experiment with a counterintuitive strategy, where matching is in reverse order (i.e., lower student layers matched to upper teacher layers). 
This seemingly nonsensical strategy sheds light on the mechanism of intermediate-layer matching.

\textbf{Out-of-Order Random Matching.} We choose the same teacher layers as forward matching, then randomly shuffle the order. The order is maintained during distillation.
We average the performance across five seeds to evaluate the effect of different random mappings.

Note that the intermediate-layer matching loss is combined with the predictor's KL loss by $\mathcal L = \mathcal L_{\text{KL}} + \lambda \mathcal L_\text{hid}$, where $\lambda$ is a hyperparameter to balance the losses. We experimentally determined $\lambda=3$ to be a good balance between KL and intermediate-layer matching losses, since it produces improved performance over the {No Matching} ($\lambda = 0$) baseline, while higher values of $\lambda$ may negatively impact performance.

We emphasize that across all layer-selection strategies, we fix the set of teacher layers that are available for the student to learn from. In other words, the only difference among the layer-selection strategies is the \textbf{order} of matching. 

\section{Results and Analysis}
\label{sec:results}

\begin{figure}[t!]
    \centering
    \begin{subfigure}{\linewidth}
        \includegraphics[width=0.95\linewidth]{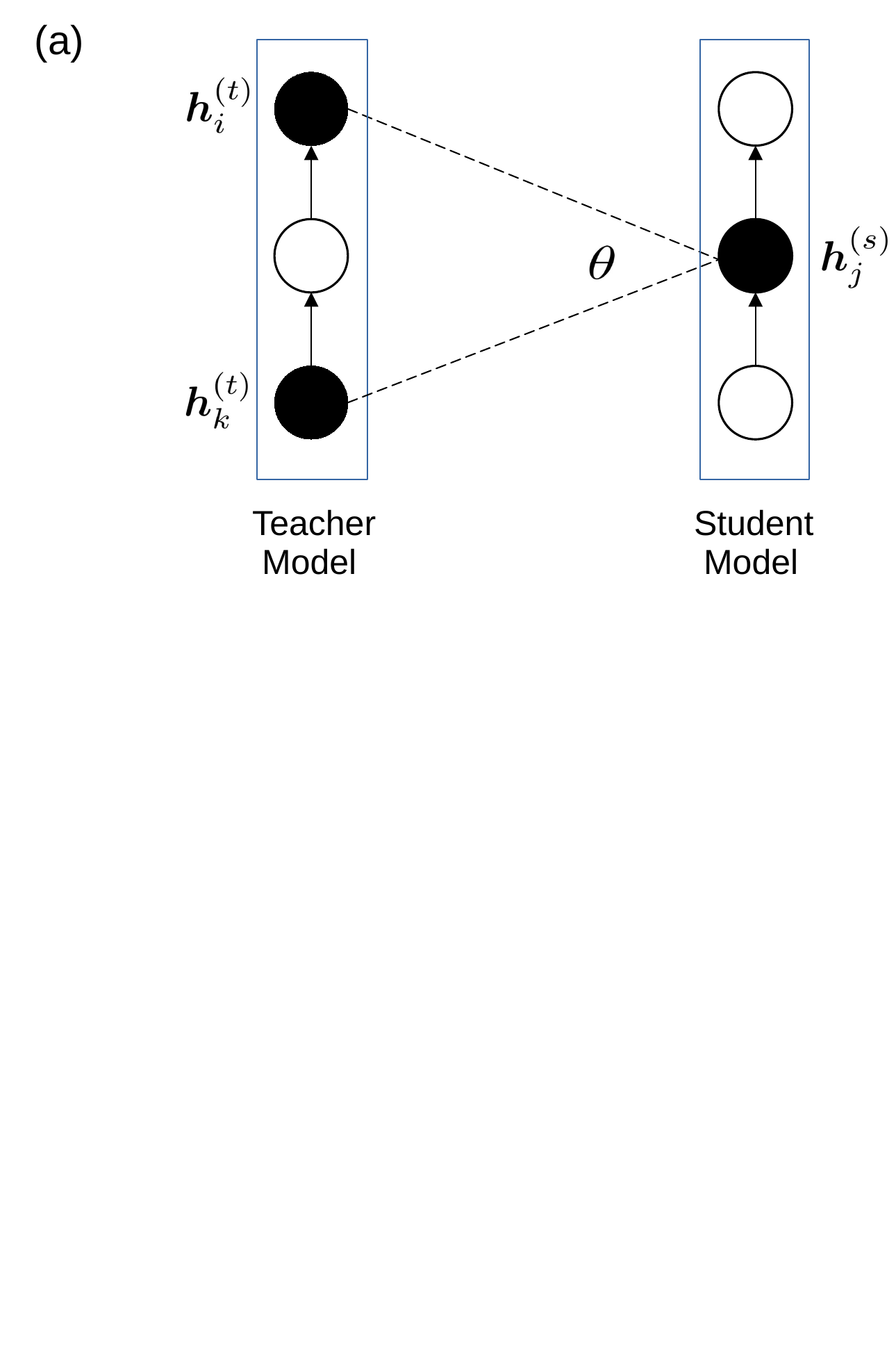}
    \end{subfigure}
    \begin{subfigure}{\linewidth}
        \includegraphics[width=0.95\linewidth]{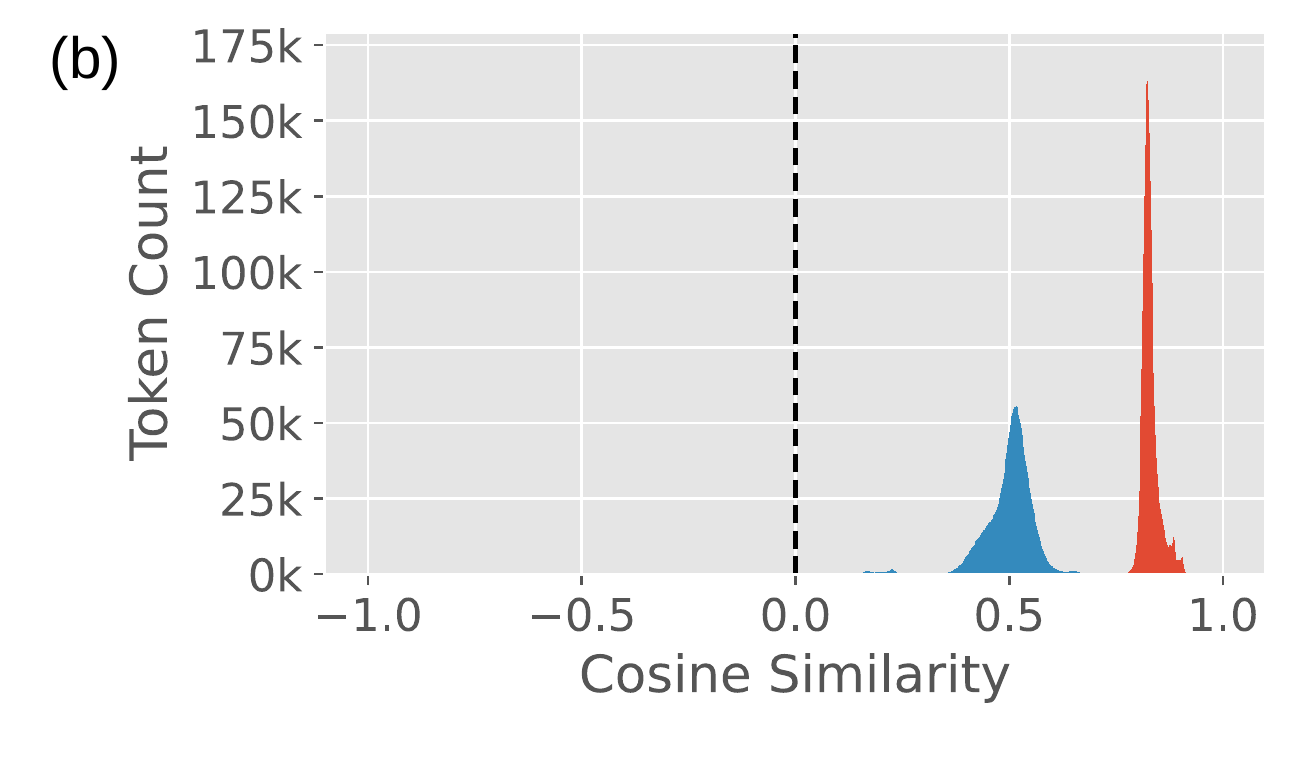}
    \end{subfigure}
    \begin{subfigure}{\linewidth}
        \includegraphics[width=0.95\linewidth]{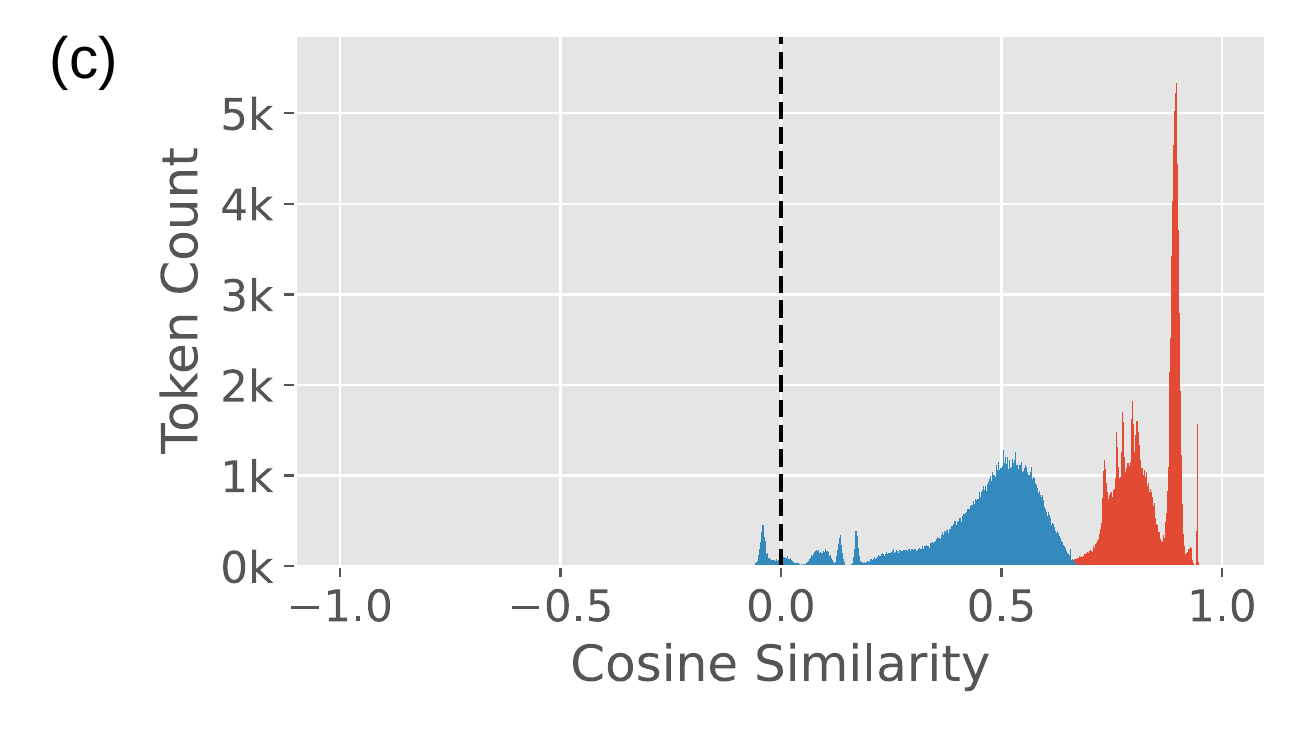}
    \end{subfigure}
    \begin{subfigure}{\linewidth}
        \includegraphics[width=0.95\linewidth]{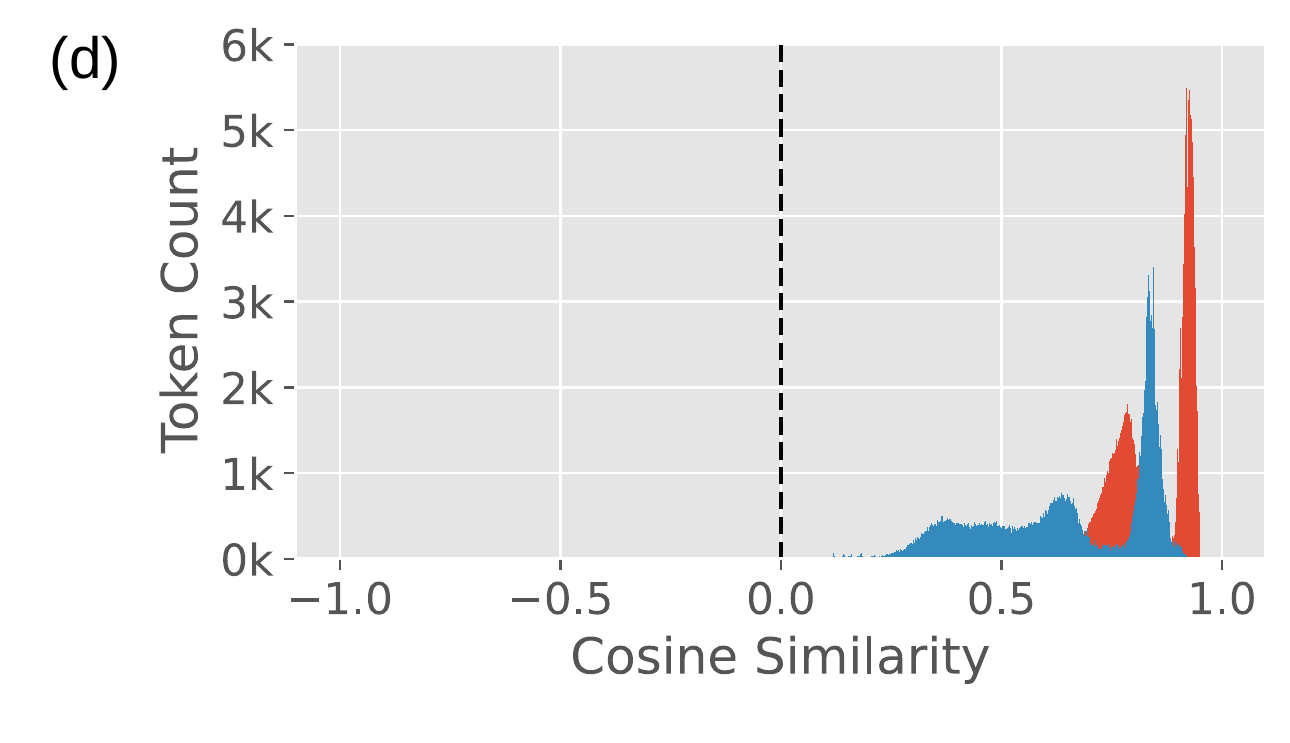}
    \end{subfigure}
    
    \caption{(a) Illustration of the angle calculation. Cosine similarities are shown for (b) MNLI classification, (c) Encoder in the  WMT task, (d) and Decoder in WMT (bottom). {\color[RGB]{226,74,51} Orange} refers to the setup of random parameter initialization and {\color[RGB]{52,138,189}blue} refers to student weights initialized by the teacher.}
    \label{fig:angles}
\end{figure}

\textbf{Setups.} 
We analyzed different layer-selection strategies on both classification and generation tasks.
For classification, we adopt the widely used MNLI~\citep{mnli}, QQP,\footnote{\href{https://www.kaggle.com/c/quora-question-pairs}{\texttt{https://www.kaggle.com/c/quora-question-pairs}}} QNLI~\citep{squad}, and SST-2~\citep{sst2} using the BERT model~\citep{bert}. We also include more challenging tasks, namely HellaSwag~\citep{hellaswag} and CommonsenseQA~\citep[CoQA;][]{commonsense}, using the more recent Qwen3 models~\citep{qwen3}.
For generation, we use the popular DART~\citep{dart} and WMT16 En--Ro~\citep{wmt16} datasets. We evaluate BART~\citep{bart} on the DART benchmark and T5~\citep{t5} on WMT16, respectively. 
Our selection of tasks and models give a comprehensive coverage of different tasks and model architectures.

For moderate-sized student models (BERT, BART, and T5), we explore two parameter initialization strategies: copying the weights from select teacher layers and random initialization. The former is a practical method used to quickly transfer knowledge from select teacher layers to the student~\citep{distilBERT,predistill}, and the latter is used to study the effects of layer-selection strategy in isolation. For the Qwen3 large language model, we only perform weight-copying because it is often not feasible to directly finetune large models on downstream tasks without warming up the weights (commonly achieved through pretraining).

More experimental details can be found in Appendix~\ref{app:datasets}.

\textbf{Main Results.} In Table~\ref{tab:tab1}, we present the main results of our layer-selection experiments. In Rows~1--2, our finetuned teachers perform similarly to previous work~\cite{bert,dart,fdistill}, showing that we have successfully set up the environment for KD experiments.

\setlength{\belowcaptionskip}{-10pt}
    
    

We examine different layer-selection strategies. As shown in Rows~9--12, the student model achieves similar results across different strategies, with only 2--3 points difference in accuracy for classification tasks and 1--2 points difference in BLEU for generation tasks. Notice that Reverse Matching and Out-of-Order Random Matching appear nonsensical, when in fact they still achieve close performance to Forward Matching, largely outperforming No Matching. The results show that layer-selection strategy has an unexpectedly small effect on student performance.

It should be emphasized that intermediate-layer matching indeed helps KD compared with No Matching,\footnote{One exception is the DART experiment with randomly initialized weights, for which we suspect intermediate-layer matching causes the student to overfit. That said, different strategies still perform similarly to conventional Forward Matching, and thus, it does not contradict our general finding.} even though the matching strategy does not play a significant role. On MNLI, for example, all strategies improve upon No Matching by six to ten points with random weight initialization and two points with weight-copying.

Next, we take a closer look at how different layer-selection strategies behave under the two parameter initialization settings. To reiterate, weight-copying is a simple and practical method of transferring the teacher's knowledge to the student~\citep{distilBERT,predistill}. We also experiment with randomly initialized students in order to disentangle the effects of layer-matching. In Rows~4--7, we see that layer-selection strategies perform similarly to one another, and for the most part better than No Matching.


\mycolor{red}{\textbf{Results on Challenging Tasks.} 
We further compare different layer-selection strategies on the more challenging HellaSwag and CoQA tasks, which involve more reasoning. As such, we employ the newer Qwen3 models.
As shown in Table~\ref{tab:llm}, we observe a similar trend as in our main results, i.e., different strategies generally produce similar results. This suggests the generality of our findings.}


\textbf{An Interpretation Based on Vector Angles.} 
A curious question arises from these observations: why does intermediate-layer matching help KD, but different layer-selection strategies perform similarly? To answer this, we measure the angles between the teacher's layers, viewed from the student. Specifically, we measure the angles formed by two teacher layers' and one student layer's vector representations, depicted in Figure~\ref{fig:angles}a. We show the phenomenon in the MNLI and WMT16 En--Ro datasets in Figures.~\ref{fig:angles}b, \ref{fig:angles}c and \ref{fig:angles}d. We see that in both randomly initialized and weight-copied settings, the cosine similarity is positive, suggesting that the angles are mostly acute. In other words, the student layer is pulled to the same general direction regardless of which teacher layer it is matched to. This finding is consistent with~\citet{shortgpt} and~\citet{ineffectiveness}, where they show that different layers may contain similar knowledge.   

\textbf{Additional Results.} We further analyze the student depth in Appendix~\ref{sec:b} and experimental stability in Appendix~\ref{app:stability}.

\section{Concluding Remarks}

In this paper, we observe an intriguing phenomenon that, although intermediate-layer matching helps knowledge distillation, the layer-selection strategy does not matter (much); we also provide an interpretation based on the angles of teacher and student layers. Our work suggests potential limitations and oversights in previous work, where researchers present various heuristic layer matching methods when training their distilled systems, but their effect is not comprehensively studied. We advise the KD practitioners to focus their efforts on other areas of KD, for example, loss functions, initializations, and representation learning.

\section{Limitations}

In our work, we have experimented with various setups, including eight tasks (six classification and two generation), four model architectures, and two parameter initialization methods. Although the results are generally consistent, there is one exception that intermediate-layer matching does not help in the DART setup. This is understandable as empirical findings are often noisy. We suspect that it is due to the student model overfitting to the teacher's representations, since we are training a wider student model (compared to T5 and BERT students) from randomly initialized weights on a small dataset. That said, this does not contradict our conclusions, as all layer-selection strategies still perform equally bad.

Additionally, we clarify that our work focuses on KD in the fine-tuning regime (for a certain task) instead of pretraining, due to the limited resources. It is noticed that fine-tuning is the typical setting of how most practitioners perform KD. We also mention that this paper is intended for an NLP audience, therefore non-NLP experiments (such as computer vision) are outside the scope of this work.

It is also worth mentioning that our work does not suggest intermediate-layer matching is unhelpful for KD. Rather, we present an interesting phenomenon that the layer-selection strategy plays an insignificant role in the process. We argue that future studies on layer selection should have a more rigorous comparison of its effect.


\section{Acknowledgments}
We thank the reviewers and area chairs for their efforts.  The research is supported in part by
the Natural Sciences and Engineering Research Council of
Canada (NSERC), the Amii Fellow Program, the Canada CIFAR AI Chair Program, a donation from DeepMind, and the Digital Research Alliance of Canada (alliancecan.ca).
\bibliography{custom}


\appendix

\begin{table*}[t!]
\
\centering

\resizebox{1.4\columnwidth}{!}{
\begin{tabular}{|l|l|l|r|ccc|}
\hline
Model & Depth & \begin{tabular}[c]{@{}l@{}}Layer Matching\end{tabular} & \multicolumn{1}{c|}{\#} & \begin{tabular}[c]{@{}c@{}}MNLI-m/mm\\ Acc\end{tabular} & \begin{tabular}[c]{@{}c@{}}DART\\ BLEU\end{tabular} & \begin{tabular}[c]{@{}c@{}}WMT16\\ BLEU\end{tabular} \\ \hline
Teacher & 12-layer & -- & 1 & 84.5 \ / \ 84.1 & 48.80 & 25.90 \\ \hline
\multirow{15}{*}{Student} & \multirow{5}{*}{3-layer} & None & 2 & 77.4 \ / \ 76.5 & 46.32 & 22.36 \\
 &  & Forward & 3 & \textbf{79.7} \ / \ \textbf{78.8} & 47.94 & \textbf{22.65} \\
 &  & Reverse & 4 & 79.2 \ / \ 78.2 & \textbf{48.45} & 21.57 \\
 &  & All-to-one & 5 & 79.4 \ / \ 78.7 & 47.10 & 21.89 \\
 &  & Out-of-order rand & 6 & 79.0 \ / \ 78.0 & 48.18 & 22.04 \\ \cline{2-7} 
 & \multirow{5}{*}{6-layer} & None & 7 & 82.1 \ / \ 81.3 & 46.88 & 24.91 \\
 &  & Forward & 8 & \textbf{83.5} \ / \ \textbf{82.9} & \textbf{48.45} & \textbf{25.00} \\
 &  & Reverse & 9 & 82.1 \ / \ 80.9 & \textbf{48.45} & 24.30 \\
 &  & All-to-one & 10 & 82.3 \ / \ 81.8 & 48.39 & 24.44 \\
 &  & Out-of-order rand & 11 & 82.3 \ / \ 81.5 & 48.03 & 24.38 \\ \cline{2-7} 
 & \multirow{5}{*}{9-layer} & None & 12 & 84.2 \ / \ 83.3 & 46.05 & \textbf{25.88} \\
 &  & Forward & 13 & 84.1 \ / \ \textbf{83.4} & 47.66 & 25.67 \\
 &  & Reverse & 14 & 83.2 \ / \ 82.4 & 47.01 & 25.11 \\
 &  & All-to-one & 15 & 83.2 \ / \ 82.5 & 46.95 & 25.43 \\
 &  & Out-of-order rand & 16 & \textbf{84.4} \ / \ 83.3 & \textbf{47.37} & 25.41 \\ \hline
\end{tabular}%
}

\caption{Performance of different layer-selection strategies on students of different depths. Student's parameters are initialized by copying the weights of the teacher.}
\label{tab:depth}

\end{table*}

\captionsetup[table]{name={Table}}
\begin{table*}[]
\bigskip
\centering
\resizebox{0.6\textwidth}{!}{%
\begin{tabular}{|ll|c|ccc|}
\hline
\multicolumn{2}{|l|}{Model} & \multicolumn{1}{l|}{Run} & \begin{tabular}[c]{@{}c@{}}MNLI-m/mm\\ Acc\end{tabular} & \begin{tabular}[c]{@{}c@{}}DART\\ BLEU\end{tabular} & \begin{tabular}[c]{@{}c@{}}WMT16\\ BLEU\end{tabular} \\ \hline
\multicolumn{1}{|l|}{\multirow{14}{*}{\begin{tabular}[c]{@{}l@{}}3-layer \\ Student\end{tabular}}} & \multirow{7}{*}{\begin{tabular}[c]{@{}l@{}}Randomly \\ Initialized\end{tabular}} & 1 & 71.2 \ / \ 71.2 & \textbf{32.44} & 16.05 \\
\multicolumn{1}{|l|}{} &  & 2 & \textbf{72.2} \ / \ \textbf{71.8} & 32.41 & 16.90 \\
\multicolumn{1}{|l|}{} &  & 3 & 70.8 \ / \ 71.1 & 33.33 & 16.95 \\
\multicolumn{1}{|l|}{} &  & 4 & 70.5 \ / \ 70.8 & 33.13 & \textbf{17.01} \\
\multicolumn{1}{|l|}{} &  & 5 & 67.9 \ / \ 67.8 & 32.35 & 16.65 \\ \cline{3-6} 
\multicolumn{1}{|l|}{} &  & \multicolumn{1}{l|}{Mean} & {\begin{tabular}[c]{@{}c@{}}70.5±1.4\\ /70.5±1.4\end{tabular}} & {32.73±0.41} & {16.71±0.35} \\ \cline{2-6} 
\multicolumn{1}{|l|}{} & \multirow{7}{*}{\begin{tabular}[c]{@{}l@{}}Weights\\ Copied\end{tabular}} & 1 & 79.3 \ / \ 78.3 & 48.18 & 21.79 \\
\multicolumn{1}{|l|}{} &  & 2 & 78.5 \ / \ 77.4 & \textbf{48.49} & 21.93 \\
\multicolumn{1}{|l|}{} &  & 3 & \textbf{79.7} \ / \ \textbf{78.6} & 47.65 & 21.86 \\
\multicolumn{1}{|l|}{} &  & 4 & 79.2 \ / \ 78.5 & 48.08 & \textbf{22.53} \\
\multicolumn{1}{|l|}{} &  & 5 & 78.5 \ / \ 77.3 & 47.54 & 21.95 \\ \cline{3-6} 
\multicolumn{1}{|l|}{} &  & \multicolumn{1}{l|}{Mean} & {\begin{tabular}[c]{@{}c@{}}79.0±0.47\\ /78.0±0.56\end{tabular}} & {47.99±0.35} & {22.01±0.27} \\ \hline
\end{tabular}%
}

\caption{Out-of-Order Random Matching experiments on MNLI, DART, and WMT16 En--Ro. For each task and parameter initialization strategy, we computed the mean and standard deviation of five runs.}
\label{tab:stddev}
\end{table*}


\section{Datasets and Models}
\label{app:datasets}
We evaluate our layer-selection strategies on a variety of classification and generation tasks.

\textbf{GLUE.} The General Language Understanding Evaluation (GLUE) benchmark is a popular suite for natural language classification. From GLUE, we chose MNLI~\cite{mnli}, QQP,\footnote{\href{https://www.kaggle.com/c/quora-question-pairs}{\texttt{https://www.kaggle.com/c/quora-question-pairs}}} QNLI~\cite{squad}, and \mbox{SST-2}~\cite{sst2}, as these tasks have large training sets and produce robust model performance. For each task, we finetune the 12-layer BERT$_\text{Base}$~\cite{bert} as the teacher. We adopted standard evaluation metrics, namely, accuracy for all tasks and $F_1$ as an additional metric for QQP.

\textbf{DART.} The DART dataset~\cite{dart} is a popular data-to-text generation task. We followed \citet{dart} and finetuned BART$_\text{Large}$~\cite{bart} with 12 encoder and 12 decoder layers, which is the teacher model in the experiment.  We report BLEU scores measuring textual overlap~\cite{bleu}.

\textbf{WMT16 En--Ro.} The WMT16 dataset~\cite{wmt16} provides parallel text between six different language pairs. For our experiments, we followed the setups in \citet{fdistill}, who chose the English--Romanian translation direction and used  100K from the 614K total samples for efficiency considerations. We also followed \citet{fdistill} and finetuned 12-layer T5$_\text{Base}$~\cite{t5} as the teacher, which has the same number of layers as the DART experiment. We also report BLEU scores as the evaluation metric.

\mycolor{red}{\textbf{CommonsenseQA.} The CommonsenseQA dataset~\citep{commonsense} is a question-answering task where each question is followed by five possible answers. We concatenated the correct answer with the question, then finetuned the 40-layer Qwen3-8B for next-token prediction to use as the teacher. Following~\citet{lm_eval}, we use zero-shot prompting to obtain responses from the language model and report exact match with the ground truth.}

\mycolor{red}{\textbf{HellaSwag.} The HellaSwag dataset~\citep{hellaswag} is a difficult sentence completion task that presents prefix sequence along with four possible endings. Like the previous task, we finetuned a 40-layer Qwen3-8B model for next-token prediction. We follow~\citet{lm_eval} and use the language model to rank each ending according to perplexity, conditioned on the prefix.}

For the student, we adopted the teacher's architecture but reduced the number of layers to three in the main experiments. Specifically, we use teacher Layers 4, 8, and 12 for matching. Note that, for BART and T5 models, this means three layers for the encoder and another three layers for the decoder. Moreover, we employed two parameter initialization strategies for the student: randomly initializing the weights and copying the weights from the corresponding teacher layer. The former isolates the effects of intermediate-layer matching from weight copying, whereas the latter is a more practical method that yields higher performance~\cite{distilBERT, predistill}.

Regarding the more challenging experiments, namely HellaSwag and CommonsenseQA, we shrunk the student by the same proportions as in the main setup. Specifically, we distilled the 40-layer teacher to a 10-layer student, using Layers 4, 8, $\cdots$, and 40 in the teacher model for matching.

\section{Analysis of Student Depths}
\label{sec:b}


We validate our intriguing phenomenon across students with different depths. Due to the limit of computing resources, we selected MNLI as the representative classification task, but include both DART and WMT16 En--Ro generation tasks.
Specifically, we experimented with student models containing three, six, and nine layers, initialized by copying the teacher's weights.
As seen in Table~\ref{tab:depth}, different layer-selection strategies show similar performances, confirming that the layer-selection strategies do not matter (much) across student models with various depths.

\section{Experimental Stability} 
\label{app:stability}
In the main results (Table~\ref{tab:tab1}), excluding the Out-of-Order Random Matching setup, we ran every experiment only once due to the large number of models, tasks, and setups. 

We show the stability of our results in Table~\ref{tab:stddev} by computing the standard deviation of five runs. Here, we chose Out-of-Order Random Matching because this is in theory the most noisy setup due to the stochasticity of layer matching. 

In Table~\ref{tab:stddev}, we see that random initialization yields higher standard deviation than the weight-copied setting. This is understandable, as the former setup involves more randomness. Nevertheless, the model performs stably in both settings, showing that our results and findings are reliable.

\end{document}